\documentclass[sigconf]{acmart}
\AtBeginDocument{%
  }

\copyrightyear{2025}
\acmYear{2025}
\setcopyright{acmlicensed}
\acmConference[SIGIR '25] {Proceedings of the 48th International ACM SIGIR Conference on Research and Development in Information Retrieval}{ July 13--18, 2025}{Padua, Italy.}
\acmBooktitle{Proceedings of the 48th International ACM SIGIR Conference on Research and Development in Information Retrieval (SIGIR '25), July 13--18, 2025, Padua, Italy}
\acmDOI{10.1145/3726302.3729988}
\acmISBN{979-8-4007-1592-1/2025/07}

\usepackage{subfigure}
\usepackage{adjustbox} % 核心包
\usepackage{graphicx}  % 图形处理
\usepackage{xcolor}    % 颜色支持
\usepackage{pifont}
\begin{document}

\title{Generative Meta-Learning for Zero-Shot Relation Triplet Extraction}

\author{Wanli Li}
\orcid{0000-0003-0670-5397}
\affiliation{%
  \institution{Huazhong Agricultural University}
  \city{Wuhan}
  \state{Hubei}
  \country{China}
}
\email{liwanli@mail.hzau.edu.cn}

\author{Tieyun Qian}
\orcid{0000-0003-4667-5794}
\affiliation{%
  \institution{Wuhan University}
  \city{Wuhan}
  \state{Hubei}
  \country{China}
}
\email{qty@whu.edu.cn}

\author{Yi Song}
\orcid{0009-0009-0579-7972}
\affiliation{%
  \institution{Huazhong Agricultural University}
  \city{Wuhan}
  \state{Hubei}
  \country{China}
}
\email{songee@webmail.hzau.edu.cn}

\author{Zeyu Zhang}
\authornote{Corresponding authors.}
\orcid{0000-0002-2376-6151}
\affiliation{%
  \institution{Huazhong Agricultural University}
  \city{Wuhan}
  \state{Hubei}
  \country{China}
}
\email{zhangzeyu@mail.hzau.edu.cn}

\author{Jiawei Li}
\authornotemark[1]
\orcid{0000-0001-5787-2449}
\affiliation{%
  \institution{Huazhong Agricultural University}
  \city{Wuhan}
  \state{Hubei}
  \country{China}
}
\email{lijw@mail.hzau.edu.cn}

\author{Zhuang Chen}
\orcid{0000-0002-7048-7833}
\affiliation{%
  \institution{Central South University}
  \city{Changsha}
  \state{Hunan}
  \country{China}
}
\email{zhchen18@foxmail.com}

\author{Lixin Zou}
\orcid{0000-0001-6755-871X}
\affiliation{%
  \institution{Wuhan University}
  \city{Wuhan}
  \state{Hubei}
  \country{China}
}
\email{zoulixin@whu.edu.cn}

\renewcommand{\shortauthors}{Wanli, Tieyun, Yi, Zeyu, Jiawei, Zhuang, and Lixin.}

\begin{abstract}
Zero-shot Relation Triplet Extraction (ZeroRTE) aims to extract relation triplets from texts containing unseen relation types. This capability benefits various downstream information retrieval (IR) tasks. The primary challenge lies in enabling models to generalize effectively to unseen relation categories. Existing approaches typically leverage the knowledge embedded in pre-trained language models to accomplish the generalization process.
However, these methods focus solely on fitting the training data during training, without specifically improving the model's generalization performance, resulting in limited generalization capability. For this reason, we explore the integration of bi-level optimization (BLO) with pre-trained language models for learning generalized knowledge directly from the training data, and propose a generative meta-learning framework which exploits the `learning-to-learn' ability of meta-learning to boost the generalization capability of generative models.

Specifically, we introduce a BLO approach that simultaneously addresses data fitting and generalization. This is achieved by constructing an upper-level loss to focus on generalization and a lower-level loss to ensure accurate data fitting. 
Building on this, we subsequently develop three generative meta-learning methods, each tailored to a distinct category of meta-learning. Extensive experimental results demonstrate that our framework performs well on the ZeroRTE task. Our code is available at \url{https://github.com/leeworry/TGM-MetaLearning}.
\end{abstract}





\maketitle

\section{Introduction}
\begin{figure}[t]
\center{
\includegraphics[scale=0.4]{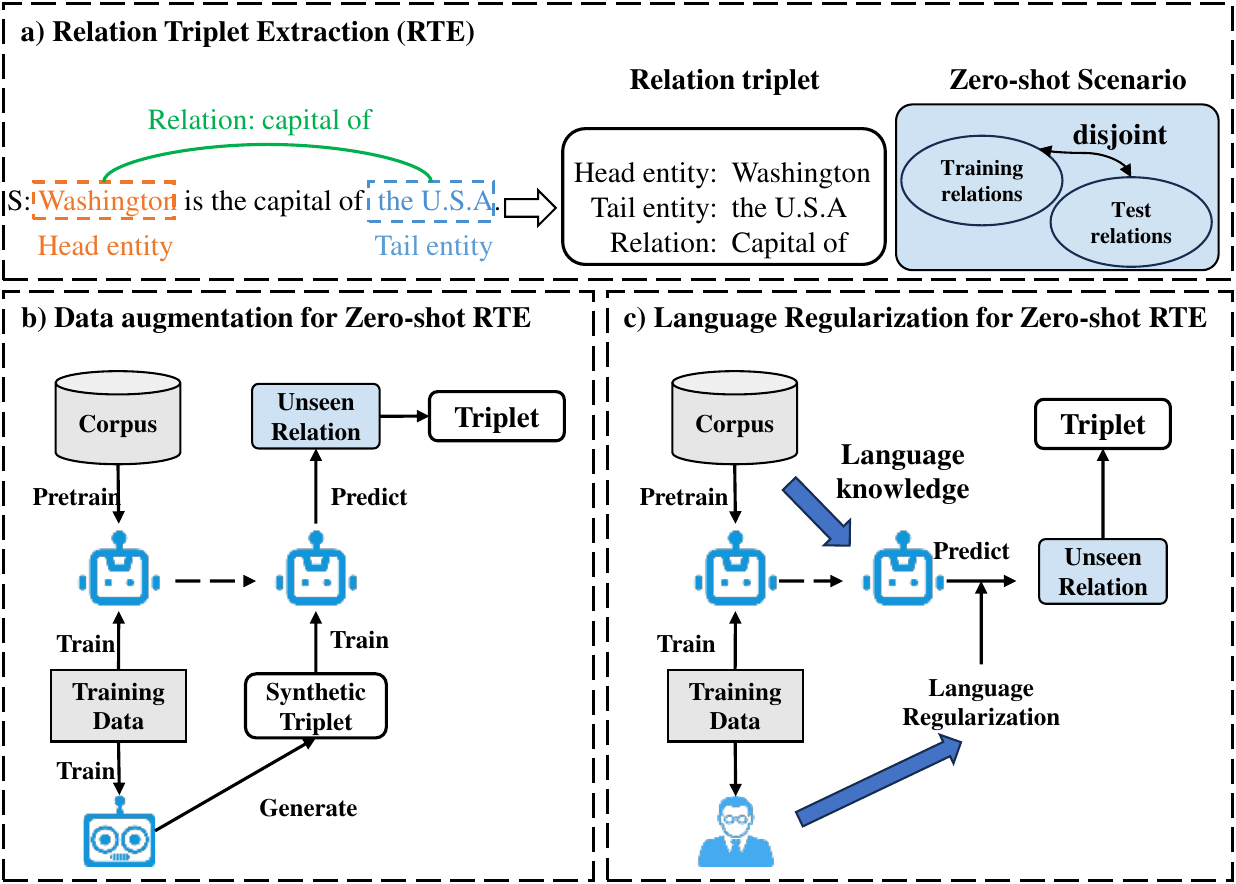}}% 0.6
\caption{An illustration of the RTE task and the difference between existing methods for ZeroRTE. }
\label{fig:summary}
%Our framework based on the model generation technique generates  different models to infer on different \emph{N}-way-\emph{K}-shot tasks.
\vspace{-0.4cm}
\end{figure}

The purpose of relation triplet extraction (RTE) is to extract triplets of the form (head entity, tail entity, relation label) from unstructured text. For example, given the sentence ``Washington is the capital of the U.S.A.'' in Fig.~\ref{fig:summary} (a), RTE aims to extract the triplet (head entity: Washington, tail entity: the U.S.A., relation: capital of). RTE can transform unstructured texts into structured knowledge, which is valuable for various downstream information retrieval IR tasks \cite{sigir23_xuming,sigir24_you}.

%or few-shot relation extraction (few-shot RE)~\cite{fewREL_1,HCRP}

Existing studies followed standard deep learning and have achieved impressive performance in supervised relation extraction (RE)~\cite{two_better_one,packed_en_re_extraction} or semi-supervised RE ~\cite{DualRE} with sufficient or limited labeled data. Their training process can be formulated in Equation \ref{deep_learning}.
\begin{equation}\label{deep_learning}
w^*_\theta=\mathop{\arg\min}\limits_{w \in \mathop{\Psi}} \frac{1}{N} \sum_{i=1}^N \ell(w(x_i),y_i),
\end{equation}
where $w^*_\theta$ is the optimal weight vector $w$, parameterized by $\theta$, that minimizes the loss function, $x_i$ and $y_i$ is the input feature vector and corresponding golden label for $i$-th sample separately, and $\ell$ can be cross-entropy loss for classification, or negative log-likelihood (NLL) for language models.
Although successful in many areas, standard deep learning methods often lack sufficient data in real-world scenarios. This scarcity limits the training of models on less common relations, making the enhancement of model generalization to handle unseen relations a critical research challenge.

%Standard deep learning methods have achieved certain successes, but in real-world scenarios, they often fail to meet the data quantity requirements. Relation categories in real-world scenarios follow a long-tail distribution, resulting a limited training data problem for those relations in tail. Consequently, addressing the model's generalization ability and enabling it to effectively handle unseen relations is a highly valuable research challenge.

%As a result, many researchers are exploring the generalization problem in relation extraction tasks, with the zero-shot RTE problem being the most challenging. Following this line, ~\citet{relationprompt} firstly introduce the ZeroRTE task to extract both the unseen relation and the corresponding entities. The task~\cite{wiki-zsl,reading-comprehension} is proposed to extract never-before-seen relations and corresponding entities based on the relation.

As a result, many researchers are focusing on the generalization problem in RE, with zero-shot RTE being the most challenging. Along these lines, \citet{relationprompt} firstly introduced the ZeroRTE task to extract unseen relations and their corresponding entities. The purpose is to extract relations and entities that have not been seen before, based on training data \cite{wiki-zsl, reading-comprehension}.

Existing ZeroRTE methods treat the generalization issue as a \emph{distribution shift problem}, operating under the assumption that the distributions of the training and test datasets are distinct. As illustrated in Fig.~\ref{fig:summary} (b, c), existing methods are generally classified into two approaches. One category involves data augmentation techniques that leverage the text generation capabilities of language models to create training data for unseen relations. The other approach involves using regularization methods to constrain model capacity and retain semantic knowledge, thus enhancing its generalization performance.

%These models adjust their parameters by leveraging the knowledge in synthetic datasets for unseen relation categories generated by language models (LMs) to adapt to the new distribution (unseen relation categories). 
Based on synthetic data of unseen relations, models can adapt to the new distribution, the process can be formulated as follows:
\begin{equation}\label{distribution_shift}
w^*_\theta=\mathop{\arg\min}\limits_{w \in \mathop{\Psi}} \frac{1}{N} \sum_{i \in D^{syn}} \ell(w(x_i),y_i),
\end{equation}
where $D^{syn}$ is the synthetic data set generated by language models. 
For example, RelationPrompt adopts a pre-trained generative model (GPT-2) to generate synthetic samples for unseen relations. KBPT uses a prompt model to synthesize training samples~\cite{KBPT}.

The data augmentation approach has two primary limitations. Firstly, the model's performance is constrained by the quality of the generated text, which can be of low quality since the text generator operates without ever having seen real samples, making the output quality unpredictable and potentially unreliable. Secondly, this method lacks efficiency. The model requires retraining to accommodate new classes, which is time-consuming \cite{continualRE}.

%However, ground-truth entities are not always readily available in real applications. If a model can generalize to unseen relation categories, its applicability across various scenarios can be significantly enhanced.
%There is only one study named RelationPrompt \cite{relationprompt} on the zero-shot RTE (ZeroRTE).
%are still unable to address the challenges of the realistic world that is constantly updating and changing. They

%In order to mitigate the constraints associated with data generation, numerous models opt to leverage the inherent pre-trained knowledge of the model by utilizing language.
Language regularization methods design constraints to force models to retain generic semantic knowledge in order to adapt to unseen relation categories.
For example, PCRED employs a knowledge-based prompt to increase performance in ZeroRTE \cite{pcred}. \citet{zett} transforms ZeroRTE into a template-filling paradigm, leveraging the model's pre-existing knowledge to effectively generalize to unseen relations. The process can be formulated as follows:
\begin{equation}\label{deep_learning_prompt}
w^*_\theta=\mathop{\arg\min}\limits_{w \in \mathop{\Psi}} \left( \frac{1}{N} \sum_{i=1}^N \ell(w(x_i, \mathcal{P}), y_i) + \lambda R(w) \right),
\end{equation}
where $w(x_i, \mathcal{P})$ denotes the model's prediction based on input data $x_i$ augmented with a prompting phrase $p$. This prompting phrase $p$ can encapsulate pre-trained knowledge, guiding the model to better leverage such information for predictions. By adjusting the content and structure of the prompting phrase, one can effectively integrate pre-trained knowledge into the model's prediction process, thereby enhancing model performance and generalization capabilities. $R(w)$ serves as a regularization term quantifying the disparity in model parameters $w$ and pre-trained knowledge, with $\lambda$ controlling the weight of the regularization term during optimization. Tuning $\lambda$ balances the model's training performance and retention of pre-trained knowledge. Since pre-trained language models contain a vast amount of knowledge unrelated to ZeroRTE, it is challenging to design appropriate regularization terms that effectively activate the relevant capabilities. Moreover, the model's generalization ability relies heavily on pre-training knowledge while neglecting the task-specific knowledge provided in the current training data, resulting in the limited generalization of the model.

%This paper argues that in addition to leveraging pre-trained knowledge, we should also capture as much generalized knowledge as possible from the training data. Therefore, the training target becomes learning relation triplet extraction and learning generalized knowledge. These two issues are intertwined, and balancing them is challenging when using only one single training objective function.
The key to improving model generalization performance is to capture as much generalized knowledge as possible from the training data in addition to leveraging pre-trained knowledge.
Therefore, the training target becomes learning RTE and learning generalized knowledge. These two issues are intertwined, and balancing them is challenging when using only one single training objective function.
In order to balance the two targets, we propose a generative meta-learning framework based on bi-level optimization (BLO) \cite{blo}.
The standard BLO problem contains two levels of optimization tasks:
\begin{equation}\label{blo_equation}
\min_{x \in \mathcal{X}}F(x,y), s.t. y\in \mathcal{S}(x),
\end{equation}
where $y\in \mathbb{R}^n$ and $x\in \mathbb{R}^m$ are respectively referred to Lower-Level (LL) and Upper-Level (UL) variables, $F$ is the UL objective, and $\mathcal{S}(x)$ is the solution of the LL subproblem.
Specifically, the UL subproblem in the context of ZeroRTE pertains to the discovery of a relational triplet extraction pattern conducive to enhancing the model's ability to generalize across novel categories, and the LL subproblem involves the comprehensive acquisition of triple extraction knowledge embedded within the training dataset.
In addition, there are various meta-learning techniques that can aid in improving generalization. Therefore, we further explored the combination of three meta-learning techniques including metric learning, gradient optimization, and model architecture adjustments.

%Within the framework of Bi-level Optimization (BLO), we firstly incorporate meta-learning techniques and formulate a diverse array of tasks aimed at capturing broadly applicable meta-knowledge. To improve the model's generalization, we further investigate three avenues for refining the language model: metric learning-based approaches, model-based strategies, and optimization-based methodologies, and achieve progress.

In summary, our contributions are as follows.
\begin{itemize}
\item This work presents the first application of BLO to enhance the generalization performance of ZeroRTE tasks, thereby establishing ZeroRTE on a novel paradigm. %Our work pioneers the utilization of bi-level optimization to enhance the generalization performance of ZeroRTE tasks. By integrating meta-learning, we elevate the model's generalization capacity, thereby establishing ZeroRTE on a novel paradigm.
\item We innovatively developed three types of generative meta-learning techniques. These advances further improve the effectiveness of the model.%We combine the concept of bi-level optimization with the pre-trained language model, introducing a prompt-based task-aware technique. This innovation facilitates the acquisition of generalized knowledge in diverse tasks. Experimental validations underscore the efficacy of this approach.
\item Detailed experimental analysis demonstrates the effectiveness of the proposed framework. Additionally, by comparing the performance of different meta-learning methods, we conclude that the design of meta-learning should be consistent with the schema of the pre-trained model. %We innovatively develope three types of generative meta-learning techniques, including metric learning, gradient optimization, and model architecture. These advances further enhance the efficacy of the model.
\end{itemize}

\section{Related Work}
\subsection{Zero-shot Relation Triplet Extraction}

The extraction of the ZeroRTE is a challenging but valuable task in RE, and is first proposed by \citet{relationprompt}. 
In ZeroRTE, the model needs to learn the general knowledge of RTE based on the training data under the known relation categories, and then extract the unseen relations and the corresponding entities.

In addressing this issue, current ZeroRTE research has evolved into two main categories. The first category typically leverages data augmentation to improve the model's generalization. For example, a method called RelationPrompt employs synthetic data of unseen relations \cite{relationprompt}. This approach utilizes pre-trained BART \cite{BART} and synthetic data derived from GPT-2 \cite{gpt2} to improve generalization specifically on unseen relations. Building on this premise, numerous methods incorporate external knowledge to enrich the quality of synthetic data pertaining to unseen relations \cite{KBPT,zs-ska}. Obviously, these methods cannot adapt to unseen relations without tuning.

Another line emphasizes stressing the incorporation of prior knowledge in the pre-trained model to improve the generalization. For instance, \citet{zett} extends ZeroRTE to a template completion task, leveraging the model's knowledge to intuitively adapt to novel relations.
However, a key limitation of these approaches lies in the neglect of optimizing model generalization throughout the training process. Therefore, this paper attempts to improve model generalization performance from a BLO perspective.

\subsection{Bi-level Optimization}
The origin of BLO can be traced to the domain of game theory and is known as Stackelberg competition \cite{blo}. BLOs are hierarchical in nature, where the feasible space of the upper-level problem is constrained by the solution set mapping graph of the lower-level problem  (i.e., the second task is embedded within the first one). 

A range of machine learning methods, such as hyper-parameter optimization, adversarial training, deep reinforcement learning, and meta-learning, involve closely interconnected sub-tasks. For instance, adversarial training comprises an upper-level objective discriminator (distinguishing real samples from generator-generated data) and a lower-level objective generator (producing samples that the discriminator cannot confidently classify as real or fake). Similarly, deep reinforcement learning includes two objectives: a policy model responsible for action decisions and a value function model evaluating the quality of policies. By employing BLO, complex tasks can be decoupled to enhance model performance.

% Bi-level optimization models typically fall into three categories: singleton, optimistic, and pessimistic BLO models. \textbf{Singleton BLO} assumes the lower-level problem always has a unique optimal solution, simplifying the problem.
% \textbf{Optimistic BLO} assumes the lower-level decision-maker consistently selects the solution most favorable to the upper level.
% \textbf{Pessimistic BLO} assumes that the lower-level decision maker may not always choose the most advantageous solution, thereby introducing greater complexity into the problem.

Our framework must account for the various scenarios in BLO problems, particularly when the lower level (LL) problem has multiple optimal solutions. It becomes crucial to determine the best solution in such cases and handle them based on an optimistic BLO assumption \cite{blo_springer}. 
%The formulation of these scenarios is as follows:

% \begin{equation}\label{blo_formulation}
% \begin{cases}
%  y^*(x):=\mathop{\arg\min}\limits_{y\in \mathcal{Y}}f(x,y), ~~~~~~Singleton\\
%  y^*(x)\in \tilde{\mathcal{S}}(x):=
% \begin{cases}
% \mathop{\arg\min}\limits_{y\in \mathcal{S}(x)}F(x,y), ~~~~~~Optimistic\\
% \mathop{\arg\max}\limits_{y\in \mathcal{S}(x)}F(x,y), ~~~~~~Pessimistic
% \end{cases}
% \end{cases},
% \end{equation}
% where $\tilde{\mathcal{S}}(x)$ represents the solution set of LL subproblems. 

% In our framework, the goal of the meta-learner is to learn how to quickly adapt to new tasks or new environments, while the base learners are responsible for learning on specific tasks. In this case, the meta-learner usually selects the base learning strategies that are most beneficial to the overall learning system so that it can achieve good performance when facing new tasks. Therefore, our method is regarded as optimistic BLO.

\subsection{Meta-learning}
Meta-learning is a subfield of machine learning that focuses on developing algorithms and models capable of learning how to learn efficiently and effectively.
Due to the ability to improve the generalization capacity of machine learning models, it has attracted great interest in recent years. Meta-learning usually consists of two modules. One captures meta-knowledge (common knowledge across tasks), and the other models task-specific knowledge learned by the base learner. The key is to find meta-knowledge in this complex process. From the perspective of BLO, meta-learning can be formulated as follows:
\begin{equation}\label{likelihood1}
\omega ^*=\mathop{\arg\min}\limits_\omega \sum_{i=1}^M l ^{meta}(D_{source}^{val(i)};w_\theta^{*(i)},\omega)
\end{equation}

\begin{equation}\label{likelihood2}
%\small
s.t. ~~~~~\theta^{*(i)}(\omega)=\mathop{\arg\min}_\theta l^{task}(D_{source}^{train(i)};w_\theta;\omega),
\end{equation}
where $l^{meta}$ and $l^{task}$ respectively refer to the UL and LL objectives, $\omega$ represents the meta knowledge that needs to be learned from different tasks, and $\theta$ represents the parameter in the model.

Conventional categorizations of meta-learning methods~\cite{gradient-based_meta-learning,ARML} can be classified to metric-based, model-based, and optimization-based methods. %where the metric-based methods are the mainstream, and model-based methods are  less able to generalize to out-of-distribution tasks than optimization-based methods.
The metric-based methods~\cite{siamese_network,matching_network,prototypical_network} aim to learn an appropriate distance metric for few-shot classification and have been successfully applied to some few-shot and zero-shot tasks \cite{SimpleFSRE,HCRP}.
The model-based methods~\cite{hypertransformer,lgm-net,generate-adapter} involve a task specification to directly generate or modulate model weights.
The optimization-based methods~\cite{LEO,maml,reptile} focus on incorporating optimization within the learning process to learn an optimized initialization of model parameters.
Meta-learning offers a promising approach to addressing various challenges, particularly in the context of generalization. However, its potential remains largely unexplored in the ZeroRTE domain. In this work, we make the first attempt to incorporate BLO and meta-learning into the ZeroRTE task. To this end, we propose a generative meta-learning framework that eliminates the need for generated data and directly learns task-specific meta-knowledge during the training process.

\section{Methodology}
\subsection{Problem Formulation}
\noindent\textbf{Definition 1}  (\textbf{RTE})
Given a piece of text $s$ = ($w_1, w_2, ..., w_l$), the RTE task aims to extract the relation triplets $T=\{t^1,t^2,...,t^o\}$ in $s$. In each $t^i$= ($\tilde{e}^i_{head}$, $\tilde{e}^i_{tail}$, $r^i$ ), $\tilde{e}^i_{head}$ and $\tilde{e}^i_{tail}$ are the head and tail entities, respectively, and $r^i\in{R}$ is the relation between $\tilde{e}^i_{head}$ and $\tilde{e}^i_{tail}$, where $R$ =\{$r_{1}$, ..., $r_{|R|}$\} is a set of predefined relations.
%\{(\tilde{e}^1_{head}, \tilde{e}^1_{tail}, r^1),...,(\tilde{e}^n_{head},\tilde{e}^n_{tail}, r^n ) \}$ ,  where $\tilde{e}^n_{head}$ and $\tilde{e}^n_{tail}$ are head and tail entity respectively, and $r^n$ is the relation between $\tilde{e}^n_{head}$ and $\tilde{e}^n_{tail}$.

\noindent\textbf{Definition 2} (\textbf{ZeroRTE}) Given a seen dataset $D_S$ and an unseen dataset $D_U$, the goal of ZeroRTE is to extract triplets $T_U$ in $D_U$ by learning knowledge from $D_S$. In the zero-shot setting, the seen relation set $R_S$ =\{$r_{1}$, ..., $r_{|n|}$\} is disjoint with the unseen relation set $R_U$ =\{$r_{n+1}$, ..., $r_{|n+m|}$\}, i.e., $R_S \cap R_u = \emptyset$, where $n$ and $m$ are the sizes of the seen and unseen relations, respectively.

\subsection{Framework Analysis}
Existing ZeroRTE tasks are usually based on various pre-trained generative language models (GLMs), such as BERT, BART, T5, etc, that are usually based on the Transformer architecture. The challenges of introducing BLO in these methods are as follows:

\textit{Challenge 1.} \textbf{What type of partition can naturally decouple ZeroRTE into upper-level (UL) and lower-level (LL) sub-problems, facilitating subsequent model optimization?} 
In the intricate  ZeroRTE tasks, the shared meta-knowledge manifests in the unseen tasks, necessitating the UL task to grasp inter-task knowledge, while the LL task focuses on acquiring task-specific insights. This paper approaches model design from the task level, randomly generating a large number of tasks from the training set. LL tasks learn knowledge specific to individual tasks, while UL tasks are responsible for capturing generalizable patterns across tasks. In this process, the model needs to make inferences based on the tasks. Therefore, to effectively capture the input differences between tasks, we design a task-aware generative model.

\textit{Challenge 2.} \textbf{Can the application of different techniques (metric-based, model-based, and optimization-based) on the Transformer architecture further enhance the model's generalization performance?}
In addition to BLO, several supplementary methods can also enhance the model's capacity to effectively capture cross-task meta-knowledge. Existing meta-learning studies demonstrate that metric-based approaches, model-based techniques, and optimization-based strategies all contribute to enhancing the model's generalization capabilities. How to smoothly combine these modules with the existing knowledge in the pre-trained model is a challenge. In order to compare the impact of integrating these methods in detail, this paper redesigns the recursive generation process of language models, introduces new processes (metrics, models, optimization), and strives to further improve the generalization potential of GLMs.

%This paper is the first attempt to combine the bi-level optimization mechanism of meta-learning into the training and inference process of GLMs. 

%An overview of our proposed generative meta-learning framework is shown in Fig.~\ref{fig:main_graph}. In order to solve the problem of limited generalization capability of existing generative methods, we first develop a  generative model (TGM), which can learn the general knowledge across multiple tasks, as shown in Fig.~\ref{fig:main_graph} (a).
%We further design a complete framework for exploiting meta-learning to boost generative models. Specifically, we develop three generative meta-learning methods based on three meta-learning categories, i.e., metric-based generative meta-learning (TGM-Metric), model-based generative meta-learning (TGM-Model), and optimization-based generative meta-learning (TGM-Optimization), as shown in Fig.~\ref{fig:main_graph} (b)-(d).

\subsection{Model Overview}

\begin{figure*}[t]
\center{
\includegraphics[scale=0.54]{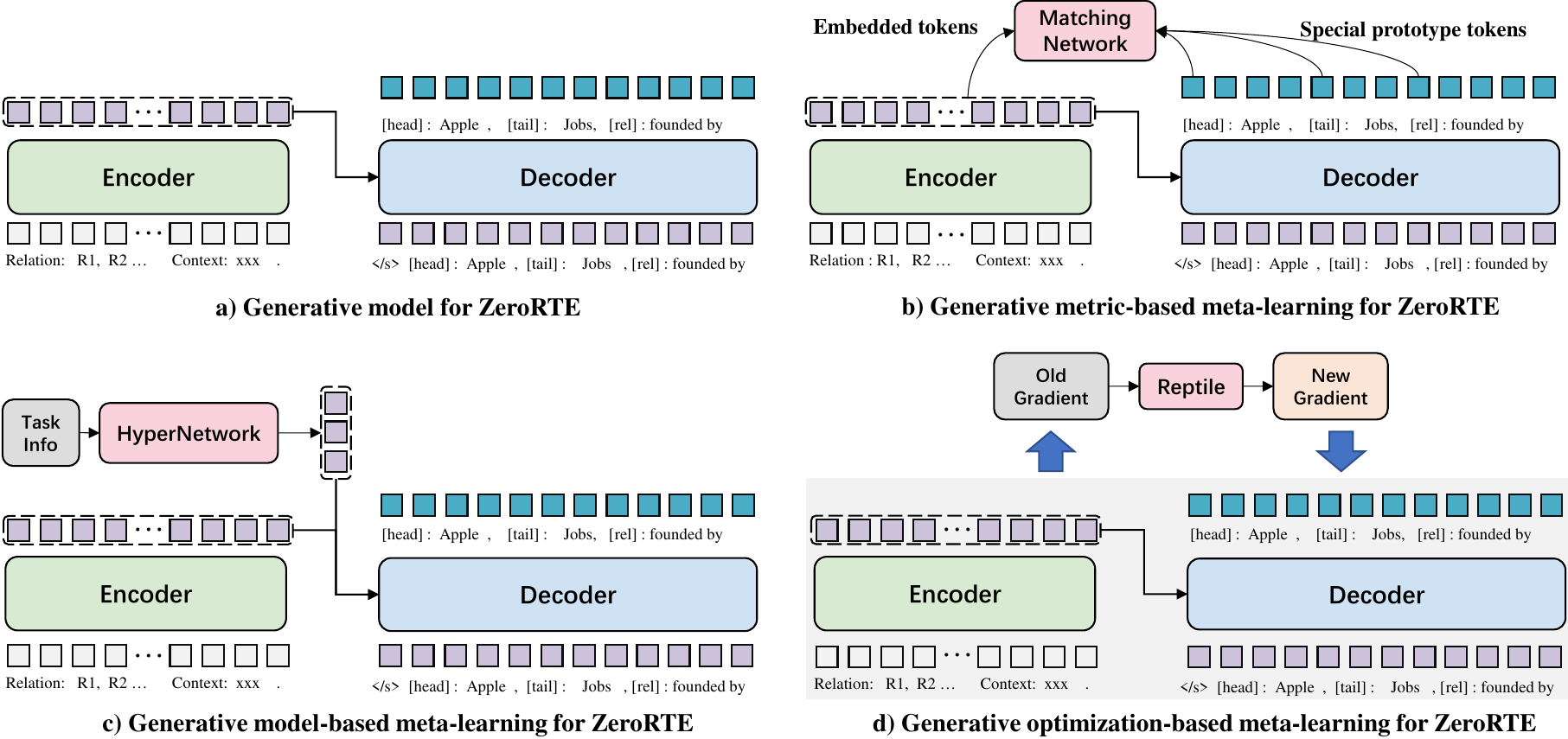}}
\caption{An overview of our proposed generative meta-learning framework for zero-shot RTE.}
\label{fig:main_graph}
\end{figure*}

An overview of our proposed generative meta-learning framework is illustrated in Fig.\ref{fig:main_graph}. To implement BLO in GLMs, we initially craft a task-aware generative model (TGM) capable of assimilating meta knowledge across diverse tasks, as shown in Fig.\ref{fig:main_graph} (a). Subsequently, we leverage different meta-learning methods for enhancing the models. Specifically, we introduce three generative metalearning methods rooted in distinct meta-learning categories: metric-based generative meta-learning (TGM-Metric) based on metric-based metalearning, model-based generative meta-learning (TGM-Model) based on model-based meta-learning, and optimization-based generative meta-learning (TGM-Optimization) based on optimization-based meta-learning, depicted in Fig.~\ref{fig:main_graph} (b)-(d).

\subsection{Task-aware Generative Model}
\label{sec:ta}
%In this subsection, we focus on developing a task-aware generative model (TGM) for the relation triplet extraction (RTE) task.
To achieve BLO within ZeroRTE, our model must possess the capability to discern between various tasks. As a result, we introduce a novel prompt for GLMs to encode task information. This approach enables the model to extract triplets corresponding to unseen relation labels at a task level rather than a sample level.

The input structure of our generative model comprises two key components. The first component entails the task information, which serves to suggest a series of potential relation categories for the given task. The second part encompasses the current text being processed. The format is as follows:
%`\textit{Relation: sitter, capital of, conflict, elector, direction. Context: Washington is the capital of the U.S.A.}'

``\textit{Relation: $R_1$, $R_2$, ..., $R_m$. Context: Washington is the capital of the U.S.A.}''

Note that $m$ is the unseen relation number, and $R_i$ is the candidate relation type in one task. For example, when $m$=5, a  task prompt might be ``\textit{Relation: sitter, capital of, conflict, elector, direction.}''.

In this way, the task-aware generative model will output the relation triplets contained in the input sentence. The predefined format is:

``\textit{Head Entity: Washington, Tail Entity: the U.S.A, Relation: capital of.}''

Notably, the task information prompt can drive the generative model to make selection among candidate relations, which is identical to the `learning to learn' idea of meta-learning. As a result,  it forms a sound basis for the subsequent meta-learning methods. Moreover, we perform multiple tasks in each epoch during training, and the optimization process is conducted across these tasks. This forces the generative model to pay more attention to the task-level information, and thus it can learn general knowledge across different tasks.

The training for the generative model is to maximize the likelihood $L(\mathcal{D})$ in the data set $\mathcal{D}$ as follows.
\begin{equation}\label{likelihood1}
%\small
L(\mathcal{D}) = \prod\limits_{i=1}^{|\mathcal{D}|}\prod\limits_{(h,t,r)\in{T_i}}P((h,t,r)|s_i,\mathcal{P}),
\end{equation}
where ($h$,$t$,$r$) refers to the (head entity, tail entity, relation), $s_i$ is the $i$-th input sentence in $\mathcal{D}$, $T_i$ is the annotated relation triplets in $s_i$, and $\mathcal{P}$ represents the task information in the input.

In bi-level perspective, TGM model contains Lower-level (LL) problem and Upper-level (UL) problem. It can be formulated as follows.
\begin{equation}\label{LL_1}
%\small
L_{\text{LL}} = \prod\limits_{i=1}^{|\mathcal{D}^{train}|}\prod\limits_{(h,t,r)\in{T_i}}P_{w_\theta}((h,t,r)|s_i,\mathcal{P}),
\end{equation}
\begin{equation}\label{UL_1}
%\small
L_{\text{UL}} = \prod\limits_{i=1}^{|\mathcal{D}^{val}|}\prod\limits_{(h,t,r)\in{T_i}}P_{w_\theta^*}((h,t,r)|s_i,\mathcal{P}),
\end{equation}
where $w^*_\theta$ represents the GLM's parameters after LL optimization. To improve the generalization, regularization terms are added to penalize the output that does not conform to the predefined format:

\begin{equation}\label{total}
%\small
L_{\text{TGM}} = L_{\text{UL}} + L_{\text{LL}} + \lambda R(w)  ,
\end{equation}

Following existing generative methods, we utilize the decoder module in Transformer~\cite{attention_is_all_you_need} to generate tokens in a recursive and sequential manner. In this way, our generative model also generates results in a predefined order. The difference is that we take the task prompt into consideration.  Given the input sentence $s_i$ and the predefined order of head entity, tail entity, and relation, the likelihood of our generative model is as follows:

\begin{align}\label{likelihood2}
\nonumber L(s_i) & = \prod\limits_{(h,t,r)\in T_i}P((h,t,r)|s_i,\mathcal{P})\\
\nonumber & =\prod\limits_{(h,t,r)\in T_i}P(h|s_i,\mathcal{P}) \cdot P(t|s_i,h,\mathcal{P})\\ &\cdot P((r|s_i,h,t,\mathcal{P})  .
\end{align}

Note that the above decoding order is experimentally explored to be optimal, and the effect of different extraction orders on the model is detailed in the  Section \ref{sec:triplet_order}.
Below we will show how TGM can be integrated into three types of meta-learning methods.

\subsection{Metric-based Generative Meta-learning (TGM-Metric)}
Metric-based meta-learning (MEML) methods learn the metric-based connections behind objects. They typically map input samples into an embedding space and then use the nearest neighbor or matching mechanism to label query samples based on the connections between their embeddings to those of labeled ones. By this means, MEML methods can naturally generalize to new domains.

Inspired by the prototype-based methods~\cite{prototypical_network,matching_network}, we design a novel feature mapping process with three prototypes output by the decoder module and a matching network for predicting whether the input tokens and the prototypes match.
Since existing generative language models follow a recursive and sequential generation manner, we insert three special tokens `[HEAD]', `[TAIL]', and `[REL]' before the relation triplet, to represent the head entity prototype, tail entity prototype, and relation prototype, respectively:

`\textit{[HEAD]: Washington, [TAIL]: the U.S.A, [REL]: capital of.}'

%To combine metric learning with the generation process without affecting the extraction of relational tripletsone

We then map the token embedding encoded by the encoder and the prototypes output by the decoder to one same vector space through a linear transformation, and then predict whether the prototypes and the corresponding token embedding by a matching network (one layer of neural network).
%The metric-based generative meta-learning model should have the ability to judge the connections between prototypes and corresponding tokens. Therefore, we design a matching network (one layer of neural network) for the generative model, so that the model has the ability to judge the relationship between the prototype and the encoded tokens.
The loss for training our TGM-Metric is the combination of the loss of the generative model and that of the matching network:
\begin{equation}
\label{metric_based_loss1}
L_{Metric} = L_{TGM} + \alpha L_{Matching}
,
\end{equation}
where $\mathcal{L}_{TGM}$ is the loss for the generative model to maximize Eq.~\ref{likelihood2} and generate the answer that conforms to the proposed format, and $L_{Matching}$ is the loss for the matching network, and $\alpha$ is a trade-off parameter.
\begin{equation}
\label{metric_based_loss2}
L_{Matching}=\sum\limits_{i=1}^{|\mathcal{D}^{train}|}\sum\limits_{j=1}^{|\mathcal{T}|}CE(G_i^j,\text{MLP}(E^j_i\oplus E^{p}_i)),
\end{equation}
where the \text{MLP} layer measures whether the $t$-th token embedding $E^t_i$ matches the $p$-th prototype embedding $E^p_i$, and $G_i^j$ is the ground-truth of matching for the 
$j$-th token in the 
$i$-th sentence with the prototype, $\oplus$ denotes the concatenation, and $CE$ is the cross-entropy loss.

\subsection{Model-based Generative Meta-learning (TGM-Model)}
Model-based meta-learning (MOML) focuses on improving the generalization ability of the model through an external module responsible for modeling meta knowledge \cite{hypernetworks,generate-adapter}. In MOML, when the external module deals with different meta-learning tasks, the module will generate optimal parameters from the perspective of the task for enhancing the inference process.

Following the MOML scheme, we explore how the external module can improve the generative models.
To this end, we first design a prompt generator module for modeling task information. We then input the newly generated parameters as encoded information into the decoder module for adapting to the generation process.
By this means, the task information and the encoder-encoded information are concatenated together and forwarded into the decoder to perform the RTE task, which forms our TGM-Model.
The training for our TGM-Model is to maximize the likelihood $L(\mathcal{D})$ in the training set $\mathcal{D}$:
\begin{equation}
\label{model-based}
L(\mathcal{D}) = \prod\limits_{i=1}^{|\mathcal{D}|}\prod\limits_{(h,t,r)\in{T_i}}\prod\limits_{j=1}^{|\mathcal{T}_i|}P((h,t,r)|s_i,\text{MLP}(\mathcal{P}_j)),
\end{equation}
where $\mathcal{P}_j$ represents the $j$-th task information prompt in the candidate task set $\mathcal{T}$ for the  $i$-th sentence.
The BLO process is consistent with the TGM model, in which the module for meta-learning is only optimized in the UL task and not in the LL task. Therefore, its formula is expressed as follows.
\begin{equation}
\label{model-based_2}
MLP_{w_\theta^{(\tau+1)}} =
\begin{cases} 
MLP_{w_\theta^{(\tau)}}- \eta \frac{\partial L}{\partial MLP_{w_\theta}}, & \text{if in UL} \\ 
MLP_{w_\theta^{(\tau)}}, & \text{if in LL}
\end{cases}.
\end{equation}

\subsection{Optimization-based Generative Meta-learning (TGM-Optimization)}
The optimization-based meta-learning (OBML) methods improve the generalization ability from the perspective of optimization and are typically model-agnostic \cite{maml,reptile}. OBML aims to find the most generalizable gradient direction among the gradients obtained by different meta-learning tasks. These methods seek to adjust the parameters of the neural network so that it can quickly adapt to different tasks.

All existing ZeroRTE methods adopt the gradient descent for training model. However, it does not consider whether the current gradient direction can improve the generalization ability or may overfit to training tasks. In contrast, our proposed task-aware generative model provides the opportunity to find the optimal gradient among different meta-learning tasks. In this subsection, we further improve our generative model with an OBML method named Reptile~\cite{reptile} which is mathematically similar to the classic MAML~\cite{reptile,maml} but is  simple
to implement. By this means, we finally get our TGM-Optimization model.
% to find the optimal gradient for zero-shot RTE.

%Frustratingly, as the first and the only one RTE method, RelationPrompt never considers whether the current gradient direction improves the generalization ability or overfits to individual samples.
%Frustratingly, it is not considered whether the current gradient direction improves the generalization ability or overfits to individual samples when training these models.
%an optimal gradient can improves the generalization ability rather than make the models overfits to individual samples.

%Existing generative model for RTE does not consider whether the current gradient direction improves the generalization ability or overfits to individual samples when training these models.
%Based on our proposed task-aware generative model, it becomes possible to find the optimal gradient from different meta-learning tasks. In this paper, we combine the generative model with an OBML named reptile that is mathematically similar to first-order MAML \cite{reptile,maml}. After that, the optimal gradient for zero-shot RTE can be found.

Formally, for an input sentence $s_i$, the likelihood in our TGM-Optimization is as follows:

\begin{align}
\label{likelihood4}
L(s_i) & = \prod\limits_{i=1}^{|\mathcal{D}|}\prod\limits_{(h,t,r)\in T_i}\prod\limits_{j=1}^{|\mathcal{T}|}P((h,t,r)|s_i,\mathcal{P}_j),%\\
% & =\prod\limits_{Task}\prod\limits_{(h,t,r)\in T_i}P(h|s_i)P(t|s_i,h)P((r|s_i,h,t)
\end{align}
where $\mathcal{P}_j$ represents the $j$-th task information prompt.

The optimization process of our TGM-Optimization model can be defined as follows:
\begin{align}
\label{optimization}
\Psi \leftarrow \Psi + \epsilon \frac{1}{n} \sum\limits_{i=1}^n(\Tilde{\Psi_i}-\Psi),
\end{align}
where $\Tilde{\Psi_i}$ is the updated parameter space on the $i$-th task which is randomly sampled, $n$ is the task number in each iteration, $\Psi$ is the model parameter space to be optimized, and $\epsilon$ is a step-size parameter.

\subsection{Inference}
%the pre-trained generative model is the main source of general knowledge, and the design of task-awareness and meta-learning enhances the ability of the model to handle the ZeroRTE task. 
In order to preserve the general knowledge contained in the generative model as much as possible, the inference in  our framework also takes the form of the generative model. Specifically, TGM-Metric utilizes meta-knowledge derived from metric learning to guide the decoder in generating the correct output. In the TGM-Model, the decoder uses meta-knowledge output by model-based meta-learner to generate the triplets. 
TGM-Optimization directly generates RTE results during inference. This is because the optimization-based meta-learner only optimizes the gradient during training and does not affect the inference process of the generative model.

\section{Experiments}
We conduct extensive experiments to verify the effectiveness of our framework and answer the following research questions: 
\begin{itemize}
    \item \textbf{RQ1}: Can the incorporation of BLO enhance the generalization capabilities of the current model?
    \item \textbf{RQ2}: What factors influence the integration of the three meta-learning techniques with models?
    \item \textbf{RQ3}: What are the advantages of this approach compared to the large language models (LLMs)?
    \item \textbf{RQ4}: What is the time complexity of the model?
\end{itemize}

\subsection{Experimental Setup}
\textbf{Datasets} We evaluate our model on two public datasets. FewRel~\cite{fewREL_1} is a standard benchmark dataset for the few-shot RE task. Wiki-ZSL~\cite{wiki-zsl} is generated with distant supervision from Wikipedia articles and the Wikidata knowledge base~\cite{wiki-kb}.
%: FewRel~\cite{fewREL_1} and Wiki-ZSL~\cite{wiki-zsl
%\citet{relationprompt} split the datasets into disjoint relation label sets for training, validation, and testing, then fit them for the zero-setting.
%The detailed data statistics are given in the Appendix.
%\begin{comment}
The detailed data statistics are shown in Table~\ref{dataset}.
\begin{table}[h]
\setlength{\tabcolsep}{1mm}
\small
\begin{center}
\caption{Statistics for two datasets.}
\vspace{-0.4cm}
\label{dataset}
\begin{tabular}{ccccc}
\hline
\textbf{Dataset} & \textbf{\#Samples} & \textbf{\#Entities} & \textbf{\#Relations} & \textbf{Sent\_len}\\
\hline
FewRel & 54,000 & 72,954 & 80 & 24.95\\
Wiki-ZSL & 94,383 & 77,623 & 113 & 24.85\\
\hline
\end{tabular}
\vspace{-0.3cm}
\end{center}
\end{table}
%\end{comment}

\begin{table*}[t]
\small
\setlength{\tabcolsep}{1mm}
\begin{center}
\caption{Comparison results on FewRel. Except for the LLMs based method (MICRE), the best scores are in bold, and the second best ones are underlined. All results are the average scores of 5 runs with the same seeds. %''$\ddagger$`` and ''$$`` indicates the statistically significant improvements with $p<0.01$ and $p<0.05$ (one-sided paired t-test) over the best baseline, respectively.
}
\vspace{-0.3cm}
\label{FewREL}
\begin{adjustbox}{center}
\begin{tabular}{lcccccccccc}
\hline
\hline
& && m=5 &&& m=10 &&& m=15& \\
Methods & Synthetic Data & Precision & Recall & $F_1$ & Precision & Recall & $F_1$ & Precision & Recall & $F_1$ \\
\hline
(1)~~TableSequence & \ding{51} & 9.86 & 9.33 & 9.59 & 12.50 & 12.02 & 12.24 & 12.46 & 11.92 & 12.19  \\
(2)~~RelationPrompt & \ding{51} & 24.70 & 24.54 & 24.69 & 24.59 & 24.23 & 24.39 & 20.66 & 20.25 & 20.45\\
(3)~~KBPT & \ding{51} & 24.14 & 23.91 & 24.02 & 24.35 & 27.28 & 26.02 & 22.11 & 21.56 & 21.83  \\
(4)~~ZS-SKA & \ding{51} & 36.24 & 34.77 & 35.49 & \textbf{33.06} & \textbf{32.85} & \textbf{32.95} & \textbf{26.11} & 23.51 & \underline{24.74}  \\
\hline
(5)~~NoGen-BART & \ding{55} & 22.90 & 22.61 & 22.75 & 16.54 & 16.31 & 16.42 & 12.16 & 11.94 & 12.05 \\
(6)~~NoGen-T5 & \ding{55} & 25.90 & 25.58 & 25.74 & 16.42 & 16.19 & 16.30 & 13.01 & 12.75 & 12.87 \\
(7)~~ZETT & \ding{55} & 31.00 & 30.61 & 30.8 & 28.91 & 27.46 & 28.17 & 24.45 & 23.42 & 23.92\\
\hline
(8)~~TGM & \ding{55} & 36.87 & 36.43 & 36.65 & 27.16 & 26.76 & 26.95 & 23.86 & 23.40 & 23.63  \\
(9)~TGM-Metric & \ding{55} & \underline{38.32} & \underline{37.86} & \underline{38.09} & 28.66 & 28.26 & 28.46 & 24.49 & 24.02 & 24.25  \\
(10)~TGM-Model & \ding{55} & 38.16 & 37.70 & 37.93 & 27.98 & 27.57 & 27.77 & 24.38 & \underline{23.89} & 24.13  \\
(11)~TGM-Optimization & \ding{55} & \textbf{39.40} & \textbf{38.91} & \textbf{39.15} & \underline{30.18} & \underline{29.77} & \underline{29.97} & \underline{25.43} & \textbf{24.94} & \textbf{25.19} \\
\hline
%(12)~~MICRE-ACC & \ding{55} &37.53 & 37.53 & 37.53 & 34.77 & 34.77 & 34.77 & 32.42 & 32.42 & 32.42\\
%\hline
\hline
\end{tabular}
\end{adjustbox}
\end{center}
\vspace{-0.3cm}
\end{table*}

\noindent\textbf{Experimental Settings}
To  make a fair comparison, we directly leverage the data split provided by RelationPrompt~\cite{relationprompt}\footnote{https://github.com/declare-lab/RelationPrompt}. In this setting, 5 random seeds are selected for the label selection process, where 5 validation labels from the seen labels are used to select sentences for early stopping and hyperparameter tuning, $m$ unseen labels ($m\in \{5,10,15\}$) are selected for testing, and the remaining sentences are treated as the training samples.

We use T5-base \cite{T5} as our pre-trained generative language model. The learning rates of the generative model parameters and other parameters are set to 3e-5 and 6e-4, respectively, and the batch size for training is set to 16. We randomly generate $t$  different tasks for each sample to improve the model's ability to capture connections between tasks and samples. For example, when $m=3$ and $t=2$, we will have two task prompts like $R_{i1}$$R_{i2}$$R_{i3}$ and $R_{j1}$$R_{j2}$$R_{j3}$. 
%Under our experimental setting, we do not distinguish whether samples within the dataset contain a single triplet or multiple triplets.

In RelationPrompt~\cite{relationprompt}, the test data is divided into two parts according to whether the sample contains a single triplet or multiple triplets. This is unrealistic in practice because we cannot know the number of triplets contained in each sample in advance.
Therefore, in our setting, the number of triplets is unknown, and the model needs to actively decode multiple triplets during testing. This means that our evaluation setup is more critical than RelationPrompt.%\footnote{We also present the comparison in the appendix with the same settings of RelationPrompt.}

\noindent\textbf{Metrics} We utilize the Micro-$F1$ score (F1) as the evaluation metric, additionally reporting precision and recall for more analysis. All results are averaged across five data folds with the same seeds.

\subsection{Baseline Methods}
%ZeroRTE is a newly proposed task, and there is only one approach RelationPrompt~\cite{relationprompt} for it. Following RelationPrompt, we adapt the RTE method  \textbf{TableSequence}~\cite{two_better_one} to ZeroRTE by training on synthetic data generated by RelationPrompt. We also adopt two versions in RelationPrompt as our baselines.  One is ``\textbf{NoGen}'' which fine-tunes the BART model only on the training data. The other is the complete "\textbf{RelationPrompt}" which fine-tunes BART  using the training set and synthetic data with unseen categories generated by a fine-tuned GPT-2~\cite{gpt2}.

The compared baselines are listed as follows:
1) \textbf{TableSequence}~\cite{two_better_one}
Following RelationPrompt, we adapt the RTE method   to ZeroRTE by training on synthetic data generated by RelationPrompt. 
2) \textbf{RelationPrompt} \cite{relationprompt} is trained on the training set and synthetic data generated by a fine-tuned GPT-2~\cite{gpt2}.
3) \textbf{KBPT} \cite{KBPT} incorporates prior knowledge from ontological schemas and is trained on synthetic data generated by a generative prompt model.
4) \textbf{ZS-SKA} \cite{zs-ska} also employs data augmentation through a word-level sentence translation.
5) \textbf{NoGen-BART} is based on RelationPrompt and fine-tunes the BART model only on the training data without synthetic data.
6) \textbf{NoGen-T5} is based on RelationPrompt but fine-tunes the T5 model only on the training data without synthetic data.
7) \textbf{ZETT} \cite{zett} treats zero-shot relational triplet extraction as a template filling task and employs ranking methods to extract the relation triplet.

Within these methods, approaches 1-4 leverage data augmentation strategies to calibrate the model using synthetic data associated with unseen relations. These methods necessitate retraining when applied to unseen relations and lack natural generalization into novel domains. On the other hand, methods 5-7 do not rely on synthetic data and can be directly utilized when encountering new relations.

``\textbf{NoGen-BART}'' and ``\textbf{NoGen-T5}'' respectively fine-tunes the BART-base and T5-base on the training data. It is notable that the parameters of BART-base, GPT-2, and T5-base are 140M, 124M, and 220M, respectively. 
%ZeroRTE is able to test the ability of the model to generalize to unknown relations. The first baseline is a basic triplet extraction method named \textbf{TableSequence} \cite{two_better_one}. The method is supervised by synthetic data generated by RelationPrompt, which can then handle the ZeroRTE task. The second and fourth baselines are different versions of RelationPrompt \cite{relationprompt}. The first version of RelationPrompt is named "\textbf{NoGen}". In this version, the method finetunes the BART model only on the training data, without externally generated data. The complete "\textbf{RelationPrompt}" fine-tunes the BART model using the training set and synthetic data with unseen categories generated by a fine-tuned GPT-2\cite{gpt2}. The parameters of BART and GPT-2 are 140M and 124M respectively. To more intuitively compare the improvements brought by meta-learning methods, we added a new baseline named "\textbf{NoGen-T5}", which replaces the BART with 140M parameters in the original "NoGen" method with the T5 model with 220M parameters.

We use the source code provided by the authors of TableSequence~\footnote{https://github.com/LorrinWWW/two-are-better-than-one} and RelationPrompt
~\footnote{https://github.com/declare-lab/RelationPrompt}. We re-train them using the optimal hyper-parameters reported in their original papers. 

%\noindent\textbf{ZeroRC}
%We compared our optimization-based generative meta-learning with 4 competing ZeroRC methods. \textbf{R-BERT} \cite{r-bert} utilizes the sentence representations and perform nearest neighbor search over label embeddings. \textbf{CIM} \cite{CIM}  is an entailmentbased method which takes the sentence and each possible relation as input to perform binary classification whether the label matches the sentence semantically. \textbf{ZS-BERT} \cite{zs-bert} generates sentence representations that are conditioned on the provided entity pair information, and performs nearest neighbor search over embeddings of the candidate relation descriptions.

\subsection{Main Results}
We present the comparative outcomes of our proposed models on FewRel and Wiki-ZSL in Table~\ref{FewREL} and Table~\ref{Wiki-ZSL} correspondingly. We draw observations based on these results.

To address the question \textbf{RQ1}, we compare two models with the same structure, TGM and NoGen-T5. As shown in Table~\ref{FewREL} and Table~\ref{Wiki-ZSL}, the proposed TGM outperforms NoGen-T5 in all settings on both datasets. Given that the NoGen-T5 model shares the exact same architecture with the proposed TGM model, its superior performance clearly indicates that the BLO process facilitates the GLMs capable of capturing valid and generalizable knowledge across tasks. This finding also suggests that decomposing a complex task like ZeroRTE into UL and LL subtasks and simultaneously optimizing them can significantly enhance generalization performance of the models.

%Since the structure of the NoGen-T5 model is exactly the same as that of the TGM model, this clearly shows that the BLO process can help generative language models to capture valid and generalizable knowledge across tasks. 

\begin{table*}[t]
\small
\setlength{\tabcolsep}{1mm}
\begin{center}
\caption{Comparison results on Wiki-ZSL. Except for the LLMs based method (MICRE), the best scores are in bold, and the second best ones are underlined. All results are the average scores of 5 runs with the same seeds.}
\vspace{-0.3cm}
\label{Wiki-ZSL}
\begin{adjustbox}{center}
\begin{tabular}{lcccccccccc}
\hline
\hline
&&& m=5 &&& m=10 &&& m=15& \\
Methods & Synthetic Data & Precision & Recall & $F_1$ & Precision & Recall & $F_1$ & Precision & Recall & $F_1$ \\
\hline
(1)~~TableSequence & \ding{51} & 15.51 & 11.86 & 13.46 & 10.64 & 5.51 & 8.07 & 8.67 & 5.55 & 7.08  \\
(2)~~RelationPrompt & \ding{51} & 24.91 & 20.46 & 22.39 & 19.27 & 16.19 & 17.57 & 14.20 & 12.31 & 13.48\\
(3)~~KBPT & \ding{51} & 35.45 & 31.64 & 33.50 & 22.41 & 21.74 & 23.57 & 21.02 & 17.31 & 19.01\\
(4)~~ZS-SKA & \ding{51} & \textbf{45.27} & \textbf{41.68} & \textbf{43.40} & \textbf{29.88} & \textbf{26.05} & \textbf{27.83} & \textbf{23.67} & \textbf{20.39} & \textbf{21.93}\\
\hline
(5)~~NoGen & \ding{55} & 18.81 & 15.41 & 16.87 & 11.94 & 10.16 & 10.96 & 8.43 & 6.95 & 7.62  \\
(6)~~NoGen-T5 & \ding{55} & 19.21 & 16.52 & 17.66 & 12.25 & 10.33 & 11.19 & 9.05 & 7.48 & 8.19 \\
(7)~~ZETT & \ding{55} & 26.22 & 23.76 & 24.93 & 21.05 & 18.99 & 19.97 & 18.31 & 13.99 & 15.86\\
\hline
(8)~~TGM & \ding{55} & 34.40 & 27.76 & 30.64 & 22.02 & 18.54 & 20.10 & 16.93 & 14.03 & 15.34  \\
(9)~TGM-Metric & \ding{55} &  37.35 & 30.17 & 33.25 & 24.77 & 20.96 & 22.54 & 20.42 & 16.86 & 18.47 \\
(10)~TGM-Model & \ding{55} & 36.72 & 30.31 & 33.51 & 24.09 & 20.84 & 22.23 & 20.10 & 16.27 & 17.99  \\
(11)~TGM-Optimization & \ding{55} & \underline{40.67} & \underline{33.42} & \underline{36.56} & \underline{26.09} & \underline{21.84} & \underline{23.73} & \underline{22.10} & \underline{18.27} & \underline{19.99} \\
\hline
%(12)~~MICRE-ACC & \ding{55} & 27.77 & 27.77 & 27.77 & 24.31 & 24.31 & 24.31 & 22.32 & 22.32 & 22.32\\
%\hline
\hline
\end{tabular}
\end{adjustbox}
\end{center}
\vspace{-0.3cm}
\end{table*}

To answer \textbf{RQ2}, we compare three generative meta-learning methods with TGM. Our proposed three generative meta-learning methods can further improve the performance of the TGM model, indicating that the meta-learning mechanism further boosts the generalization capability of the basic generative model.
After in-depth analysis, we found that the TGM-Metric and TGM-Model introduced new metric-learning modes and meta-learning modules. Although these additions can be integrated with existing semantic knowledge in Transformer, they damaged the semantic inference ability of the model to a certain extent; while the TGM-Optimization completely relies on the BLO mechanism to adjust the gradient direction and does not introduce additional features for model generalization, and it is more consistent with the original semantic capabilities of the pre-trained models. Therefore, we believe that the optimization process should be designed with careful consideration of how well it aligns with the model's original knowledge. This alignment directly influences the robustness of the model's generalization ability.

In order to analyze whether the stronger generalization of TGM-Metric comes from the influence of pre-training knowledge, we retain the structure of the T5-base model during training but reinitialize the model parameters. The experimental results show that the accuracy of all models tend to 0. This means that a large part of the model's capabilities come from pre-training knowledge. 

\begin{table}[t]
\small
\begin{center}
\caption{Comparison results with LLMs on $m=5$.}
\vspace{-0.3cm}
\label{Comparison_LLMs}
\begin{tabular}{lccc}
\hline
\hline
& \textbf{Acc (FewRel)} & \textbf{Acc (Wiki-ZSL)} & \textbf{Avg.}\\
\hline
\textbf{MICRE(LLaMA)} & 37.53& 27.77 & 32.65\\
\textbf{GPT-4o} & 11.71& 7.35 & 9.53\\
\textbf{TGM-Optimization} & 38.90  & 36.23 & 37.57\\
\hline
\hline
\end{tabular}
\end{center}
\vspace{-0.3cm}
\end{table}

\subsection{Comparison with LLMs}

%Existing LLMs have shown good generalization capabilities on a wide range of natural language processing tasks. 
For \textbf{RQ3}, we analyze the outputs of \textbf{MICRE} and \textbf{GPT-4o}. MICRE \cite{micre} based on LLaMA \cite{llama} employs in-context learning technique to tackle ZeroRTE.  We also show the accuracy of GPT-4o in the Table~\ref{Comparison_LLMs}.

 It can be seen that a complex system (GPT-4o) will produce uncontrollable and hallucinatory answers when dealing with ZeroRTE. Although we have explicitly restricted the output format of GPT-4o in the prompt, GPT-4o always outputs some extra symbols or sentences. Perhaps, simple models that learn straightforward patterns offer a more efficient way to achieving intelligence.

\begin{figure*}[t]
\center{
\includegraphics[scale=0.45]{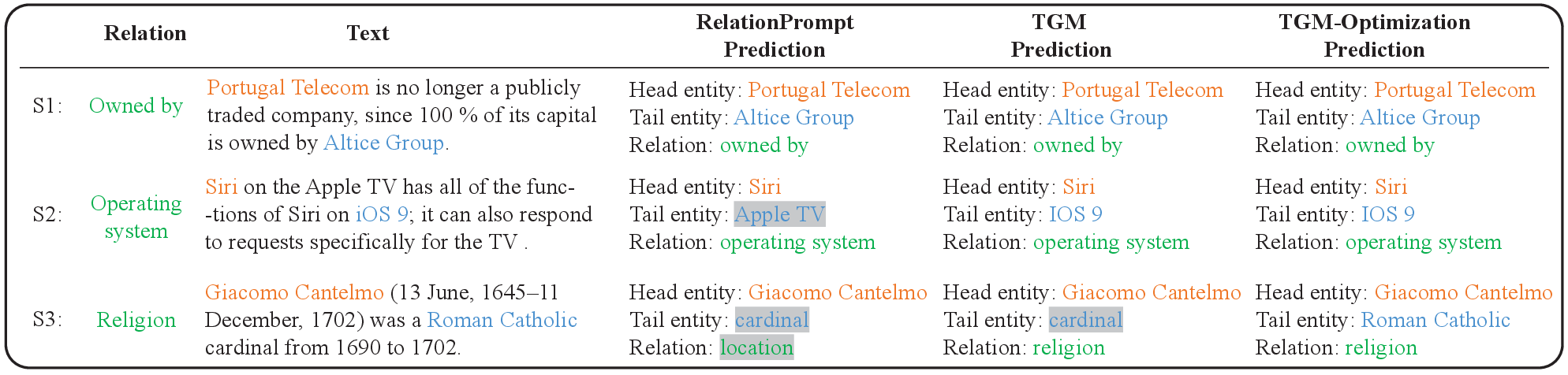}}
\caption{Case study. The orange, blue, and green tokens respectively denote the head entity, tail entity, and relation. Incorrectly extracted tokens are marked in grey.}
\vspace{-0.3cm}
\label{fig:case_study}
\end{figure*}

\subsection{Complexity Analysis}
For \textbf{RQ4}, we conduct a complexity analysis.  Both our proposed method and existing baselines utilize the pre-trained GLMs $M$ based on the Transformer structure. The computational complexity primarily depends on the backbone model $M$, which includes multi-head self-attention, feed-forward networks, layer normalization, and residual connections in each of its 
$n$ layers.\

The computation of $l$-th layer feed-forward network is formulated as $X^{L+1}=\sigma(X^lW^l)$. Where $\sigma()$ is a non-linear activation function, and $W^l$ is a feature transformation matrix $\in \mathbb{R}^{F_l\times F_{l+1}}$. For simplicity, we assume the features at every layer are size-$d$. As such, $W^l$ is an $d\times d$ matrix. From the setting, we know $d_v=d$.

We analyze the time complexity of the TGM by three high-level operations:

i. Multi-head Self-Attention ($Softmax(\frac{QK^T}{\sqrt{d}})V$). We assume that it has $v$ head, then the time complexity is $O(vn^2d)$

ii. Feed-forward Network ($\sigma(X^lW^l)$) is a dense matrix multiplication between matrices of size $n\times d$ and $d\times d$. The time complexity is $O(nd^2)$.

iii. layer normalization and residual connections ($LayerNorm(X+MultiHeadAttention(X))$ and $LayerNor(X+Feedforward(X))$) involves computing mean and variance across features, which has a time complexity of $O(nd)$.
Residual connections contains a simple element-wise addition and thus has a time complexity of $O(n)$

Thus, the time complexity of TGM is $O(vn^2d+nd^2+nd+n)$ for one forward propagation, which simplifies to $O(n^2d)$ given $d<<n$.%The time complexity of backpropagation is usually the same as that of forward propagation because it requires the gradient to be computed and the parameters to be updated is the same as forward propagation. 
%We know that $d<<n$, hence the complexity can be simplified to $O(n^2d)$.
Both TGM-Metric and TGM-Model extend TGM with new feed-forward neural networks, leading to a complexity of $O(vn^2d+nd^2+nd+n+nd^2)$, still simplified to $O(n^2d)$. %After all, it still can be simplified to $O(n^2d)$.
In TGM-Optimization, we introduce a novel gradient update pattern that does not alter the model structure or inference process, maintaining a complexity of $O(n^2d)$.
%In TGM-Optimization, we designed a new gradient update pattern that does not affect the model structure and inference process, so the complexity of TGM-Optimization should be $O(vn^2d+nd^2+nd+n)$. %that can be simplified to $O(n^2d)$.

In summary, the computational complexity of the proposed framework aligns with that of other frameworks.

\subsection{Impacts of the Relation Number in Prompt}
\label{sec:analysis_prompt}
Our prompt contains $m$ candidate task information in the prompt to facilitate the model to understand tasks. Following RelationPrompt~\cite{relationprompt}, we assume $m$ is known in training. However, this is impractical in real-world scenarios.
Hence we investigate whether our model can cope with an unknown number of candidate relations. For this, we set the candidate relation number as $r$, and vary it among \{2, 5, 10, 15\}. The results are shown in Table~\ref{Analysis_prompt}.
%and whether it can learn general knowledge from the task prompt with arbitrary length.
%1) As shown in Table.~\ref{Analysis_prompt}, $r$ does not affect the performance, which indicates that our model does learn general knowledge across tasks.
It is clear that a $r$ smaller than $m$ often results in a better performance~\footnote{This infers that we actually do not need to set $t$=$m$. Instead, $t$ can be a relatively small number like 5, which is another appealing property of our model.}. This is natural since the model can focus  more on the true task and is not affected by redundant relations in the prompt. However, a too small $r$  like 2  will prevent the generative model  from learning general information across tasks and hurts the performance.
%This infers that we do not need to set t=m. Instead, t can be a relatively small number like 5, which is
This finding also holds true for $r$=3, 4 for $m$=5, suggesting that a pre-determined number of relations is not necessary. This flexibility represents another attractive feature of our model.

\begin{table}[ht]
\small
\begin{center}
\vspace{-0.3cm}
\caption{Impacts of the relation number $r$ in the prompt.}
\vspace{-0.3cm}
\label{Analysis_prompt}
\begin{tabular}{lccc}
\hline
\hline
\textbf{FewRel} & \textbf{F1(m=5)} & \textbf{F1(m=10)} & \textbf{F1(m=15)}\\
\hline
\textbf{r=2} & 37.38 & 26.83 & 20.34\\
\textbf{r=5} & 39.15 &31.49 & 26.08 \\
\textbf{r=10}  & 37.84 & 29.97 & 25.90 \\
\textbf{r=15}  & 36.70 & 28.72 & 25.19 \\
\hline
\hline
\end{tabular}
\end{center}
\vspace{-0.4cm}
\end{table}

%1) As shown in Table~\ref{Analysis_prompt}, the smaller the $r$ is, the better the model performs. When the $r$ is small, the model focus  more on the understanding of the task and is not influenced by the non-correct relations in the prompt.

%2) The most appropriate value of $r$ is not $m$. The model can extract the triplets without requiring a constant $r$, which indicates that our model does learn general knowledge across tasks.

% \begin{table}[h]
% \caption{Analysis on Prompt at the setting of m=5. r represents the number of relations in prompt (task info).}
% \label{Analysis_prompt}
% \small
% \begin{center}
% \begin{tabular}{lccccc}
% \hline
% Dataset & r=3 & r=4 & r=5 & r=6 & r=7\\
% \hline
% FewRel & 39.27 & 38.92 & 39.15 & 38.59 & 38.26 \\
% Wiki-ZSL  & 37.79 & 37.56  & 36.56 & 35.66 & 35.20 \\
% \hline
% \end{tabular}
% \end{center}
% \end{table}

\subsection{Impacts of the Triplet Order}
\label{sec:triplet_order}
The decoder in Transformer generates tokens in a sequential manner and the generative model needs a predefined triplet order. We assume an HTR order in Sec.~\ref{sec:ta} and compare other two different orders THR and RHT here, where  `H', `T', and `R' denotes the head entity, tail entity, and relation, respectively.  The results are shown in Table~\ref{Analysis_order}.
%In order to explore the influence of different orders in triplets on the effect of generating models, we add  two comparative experiments that have different triplet formats: (relation, head entity, tail entity) and (tail entity, head entity, relation).

\begin{table}[h]
\small
\begin{center}
\caption{Impacts of different triplet orders on $m=5$.}
\vspace{-0.3cm}
\label{Analysis_order}
\begin{tabular}{lccc}
\hline
\hline
& \textbf{F1 (FewRel)} & \textbf{F1 (Wiki-ZSL)} & \textbf{Avg.}\\
\hline
\textbf{TGM$_{\text{HTR}}$} & 36.65 & 30.64 & 33.65\\
\textbf{TGM$_{\text{THR}}$} & 37.11  & 30.08 & 33.60\\
\textbf{TGM$_{\text{RHT}}$} & 35.16 & 29.57 & 32.37\\
\hline
\hline
\end{tabular}
\end{center}
\vspace{-0.3cm}
\end{table}

It can be seen from Table~\ref{Analysis_order} that, two generative models with the entity first order HTR and THR  are better than that with the relation first order RHT. The reason is that in the ZeroRTE task, most entities are  meaningful nouns, and the extraction of entities does not change much under different relation categories. In this case, performing the relatively easy entity extraction task first can help the subsequent hard RE task.

\subsection{Analysis on Hyper-parameters}
%The improvement brought by the Task-aware mechanism is closely related to the number of tasks sampled during training. Therefore, we design a parameter experiment on the number of tasks to explore the effect on model performance. Another critical hyper-parameter $\alpha$ is the weight of $\mathcal{L}_{Matching}$ in TGM-Metric model. $\alpha$ is used for balancing the loss of $\mathcal{L}_{Gen}$ and $\mathcal{L}_{Matching}$.

\begin{figure}[ht]
\vspace{-0.3cm}
\center{
\subfigure[Impact of $t$ on FewRel]{
\includegraphics[scale=0.22]{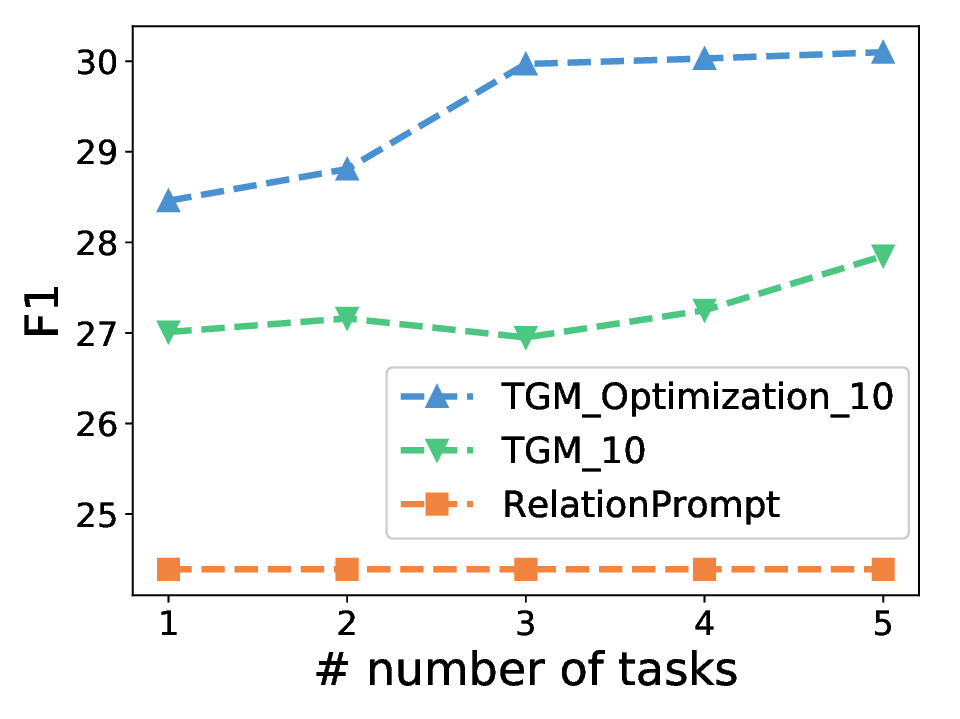}
\label{fig:task_num_fewrel}}
\subfigure[Impact of $t$ on Wiki-ZSL]{
\includegraphics[scale=0.22]{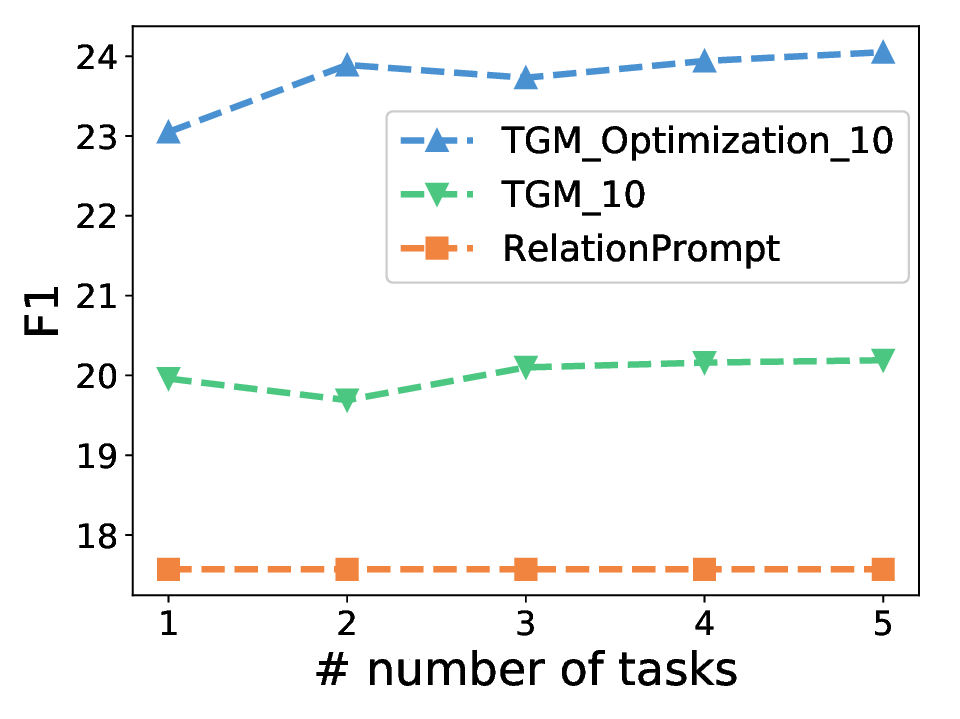}
\label{fig:task_num_wiki}}
\subfigure[Impact of $\alpha$ on FewRel]{
\includegraphics[scale=0.22]{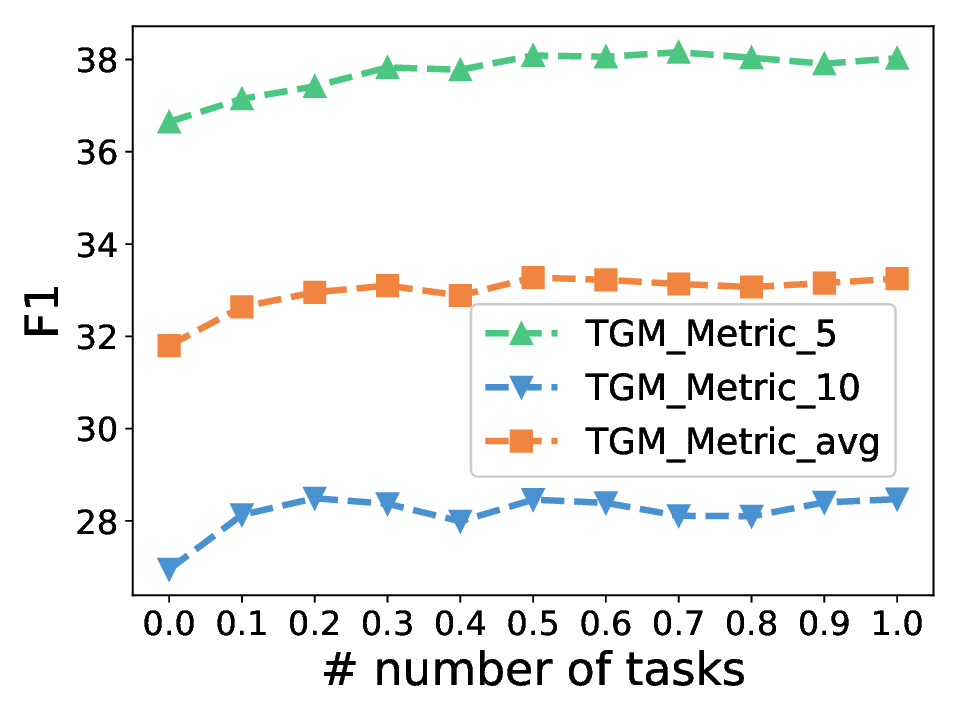}
\label{fig:alpha_fewrel}}
\subfigure[Impact of $\alpha$ on Wiki-ZSL]{
\includegraphics[scale=0.22]{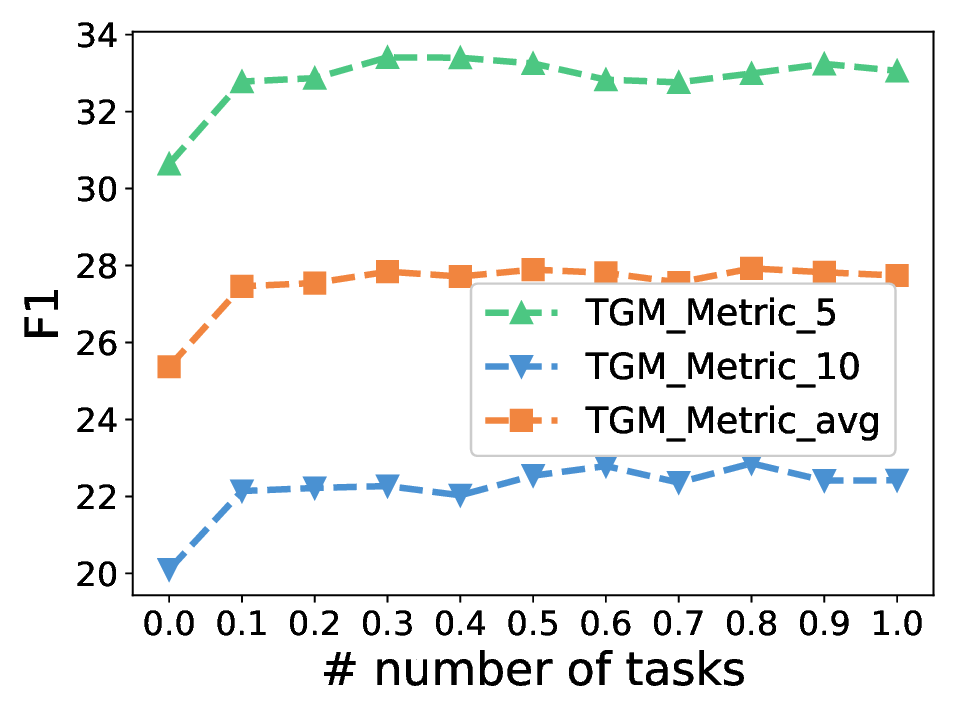}
\label{fig:alpha_wiki}}
}
\vspace{-0.3cm}
\caption{Impacts of the hyper-parameters.}
\vspace{-0.3cm}
\label{parameter}
\end{figure}

We analyse the task number $t$ and the parameter $\alpha$ for balancing the loss of meta learning and that of the GLMs in the TGM-Metric model. %$\mathcal{L}_{Gen}$ and $\mathcal{L}_{Matching}$
Figure~\ref{parameter} shows the impacts of these two hyper-parameters.

%\textcolor[rgb]{1.00,0.00,0.00}{From Fig.~\ref{fig:task_num_fewrel} and~\ref{fig:task_num_wiki}, as the number of tasks sampled by the model increases, the performance of our model tends to improve. In addition, there is an upper bound to this improvement, and the model will become stable gradually.}

From Fig.~\ref{fig:task_num_fewrel} and~\ref{fig:task_num_wiki}, we find that our two models become stable when $t$ is larger than 3. Note even when $t$=1, our models perform better than RelationPrompt since our prompt with multiple relations can provide general knowledge across tasks.
Fig.~\ref{fig:alpha_fewrel} and~\ref{fig:alpha_wiki} show that the optimal setting for $\alpha$ is about 0.5 on two datasets.

\subsection{Case Study}
%To have a close look at the impacts of the proposed task-aware prompt and meta-learning mechanism in our framework, 
As shown in Fig.~\ref{fig:case_study}, we compare the results of RelationPrompt, TGM, and TGM-Optimization on three sentences (denoted as S1, S2, and S3). 

S1 has a phrase `owned by' which explicitly points out the relation. All three generative models can extract the correct answer for such an easy task.

S2  includes the `operating system' relation. RelationPrompt predicts the correct relation and a wrong tail entity since the training of its extraction model depends on synthetic samples which may not certainly contain `Apple TV'. In contrast, both our TGM and TGM-Optimization can  extract the correct triplet. Though our models do not see samples with `operating system' relation during training, the relation name has been used as the candidate in the task prompt which will never be chosen as an answer. That is to say, the model has got the implicit knowledge that `operating system' does not refer to `Apple TV'. This demonstrates the role of generalizing knowledge across tasks.

S3 contains the `religion' relation.  RelationPrompt extracts wrong tail entity and relation and TGM model guided by the task information extracts the correct head entity  but makes mistake on the tail entity. TGM-Optimization  obtains the completely correct answer. We believe this is because TGM-Optimization retains the general knowledge about `religion' via meta-learning, which also proves the value of meta-learning.

\section{Conclusion}% add new content
In this study, we discovered that existing generative language models have limited generalization capabilities in zero-shot learning scenarios. %This limitation stems from the models' failure to utilize the inherent generalization knowledge within training data, relying instead solely on pre-trained information for enhancing their generalization performance.
To address this challenge, we propose an innovative generative meta-learning framework that exploits the synergy between BLO and multiple meta-learning strategies. Our approach effectively leverages the meta-knowledge embedded in training datasets, leading to significant improvements. %Extensive experiments conducted on two public datasets demonstrate that our framework enhances the generalization abilities of generative language models. %Additionally, we investigate various aspects including model complexity, comparisons with larger models, task sampling order, and hyperparameter optimization.

%These findings underscore the potential of integrating BLO with meta-learning as an effective strategy to boost model robustness and adaptability in previously unseen environments.

%In this study, we find that existing generative models have limited generalization capability in zero-shot learning scenarios.
%To solve this problem, 
%And we propose a novel generative meta-learning framework which exploits the potential of generative language model and meta-learning. %Specifically, we first propose a task-aware generative model which can learn the general knowledge across multiple tasks. We further develop three generative meta-learning methods which incorporate typical meta-learning schemes into the generative model to optimize the training process.
%Extensive experimental results on two public datasets  prove that our framework can improve the generalization capability of generative language models.
%and sets a new state-of-the-art performance for the ZeroRTE task.

\section{Acknowledgements}
This research project was supported in part by Hubei Key Research and Development Program of China under Grant [2024BBB055], the Fundamental Research Funds for the Central Universities, China
[Project 2662023XXQD002, 2662023XXQD003, 2662023XXQD004].

\balance
\bibliographystyle{ACM-Reference-Format}
\bibliography{sample-base}

% \appendix









\end{document}